\documentclass{article}

\PassOptionsToPackage{numbers, compress}{natbib}
\usepackage[preprint]{neurips_2026}

\usepackage[utf8]{inputenc} 
\usepackage[T1]{fontenc}    
\usepackage{hyperref}       
\usepackage{url}            
\usepackage{booktabs}       
\usepackage{amsfonts}
\usepackage{amsmath}
\usepackage{nicefrac}       
\usepackage{microtype}      
\usepackage{xcolor}         
\usepackage{algorithm}
\usepackage{algpseudocode}
\usepackage{multirow}
\usepackage{tcolorbox}
\usepackage{graphicx}
\usepackage{subcaption}   
\usepackage{booktabs} 
\tcbuselibrary{skins,breakable}
\newtcolorbox{notebox}[1]{colback=blue!5!white,colframe=blue!60!black,
  fonttitle=\bfseries\small,title=#1,left=4pt,right=4pt,top=4pt,
  bottom=4pt,fontupper=\small,breakable}
\newtcolorbox{checkbox}[1]{colback=orange!8!white,colframe=orange!70!black,
  fonttitle=\bfseries\small,title=#1,left=4pt,right=4pt,top=4pt,
  bottom=4pt,fontupper=\small,breakable}
\title{MANGO: \underline{M}eta-\underline{A}daptive \underline{N}etwork \underline{G}radient \underline{O}ptimization for Online Continual Learning}

\author{
Ankita Awasthi$^{1}$ \quad
Marco Apolinario$^{2}$ \quad
Kaushik Roy$^{1}$\\
$^{1}$Purdue University, USA \space \space $^{2}$TU Delft, Netherlands \\
\texttt{awasthi9@purdue.edu, m.apolinariolainez@tudelft.nl, kaushik@purdue.edu}
}

\newcommand{\justification}[1]{#1}
\begin{document}

\maketitle

\begin{abstract}
  In Online Continual Learning (OCL), a neural network sequentially learns from a non-stationary data stream in a single-pass with access only to a limited memory replay buffer. This contrasts sharply with off-line continual learning where training is multiple epoch dependent on large datasets. The main challenge faced by OCL is to overcome catastrophic forgetting of past tasks (stability) while learning new ones efficiently (plasticity). Existing methods counter forgetting via replay-based rehearsal, output level distillation, fixed regularization, or meta-learning on the current data. However, these methods have limitations: rehearsal introduces a stored sample bias; distillation operates on output-distributions without modulating parameter updates; fixed-regularization penalizes parameters irrespective of sensitivity; stream-only meta-learning lacks a feedback controlled parameter update. We propose Meta-Adaptive Network Gradient Optimization (MANGO), an OCL framework that balances stability-plasticity via gradient-gating and meta-learned regularization. Gradient-gating scales parameter updates based on sensitivity, preventing destructive updates. Meta-learned regularization adapts stability coefficients, evaluating the effect of parameter update on replay. In MANGO, replay acts as both a training signal and a forgetting evaluator. We evaluated our method on three standard OCL benchmark datasets. MANGO outperforms strong baselines, achieving state-of-the-art results with consistent performance across replay sizes. In domain incremental learning on CLEAR-10 and class incremental learning on CIFAR-100 and Tiny-ImageNet, it achieves highest accuracy among all baselines and achieves positive Backward Transfer, overcoming forgetting on CLEAR-10.
\end{abstract}

\section{Introduction}
Online Continual Learning requires a Neural Network  to incrementally learn from a sequential flow of non-stationary data in a strict single-pass setting,  Unlike conventional (off-line) continual learning models which are trained over multiple-passes on large, stationary datasets, OCL works on a strict online-setting, where each sample can be observed only once with a fixed memory replay of past data. Incoming data often streams one-by-one or in mini-batches, creating a computational and memory restriction compared to off-line continual learning. A model’s ability to retain knowledge from past tasks is termed as stability and the ability to efficiently learn new tasks is plasticity.  A major challenge faced by OCL models is the stability-plasticity dilemma. Learning new tasks often leads to catastrophic forgetting (CF), a phenomenon where a model is unable to retain learned knowledge from past tasks when trained on new ones. The goal of an efficient model is to balance this trade-off by learning new tasks efficiently without forgetting past samples in a single-pass.  Existing works in OCL can be categorized in four major areas for tackling CF, each with fundamental limitations. Replay-based works \citep{chaudhry2019continual, buzzega2020dark, caccia2022new} utilize rehearsal, where they maintain a small replay buffer of past samples and include them in training new tasks. While rehearsal is effective at preserving stability, it introduces a bias toward stored samples and reduces plasticity of new tasks under a small replay buffer in strict online-settings. Distillation based methods \citep{buzzega2020dark, liang2023loss} use logit matching by preserving output distributions to constrain the model behavior. This method preserves prediction stability without controlling parameter updates and their impact on past tasks. Regularization based methods \citep{kirkpatrick2017overcoming} penalize any deviation from previously learned parameters by regulating updates according to parameter importance. This approach relies on fixed heuristics which do not adapt during training according to parameter sensitivity across dynamic tasks in strict online-settings. Meta-learning based methods \citep{finn2017model, saha2025amphibian} learn to adapt gradient updates through a meta-objective, but existing approaches work on the current stream of data. They lack a feedback mechanism to reduce CF over a long sequence of tasks.

\begin{figure}[t]
\label{mango-results}
\centering
\includegraphics[width=0.95\columnwidth]{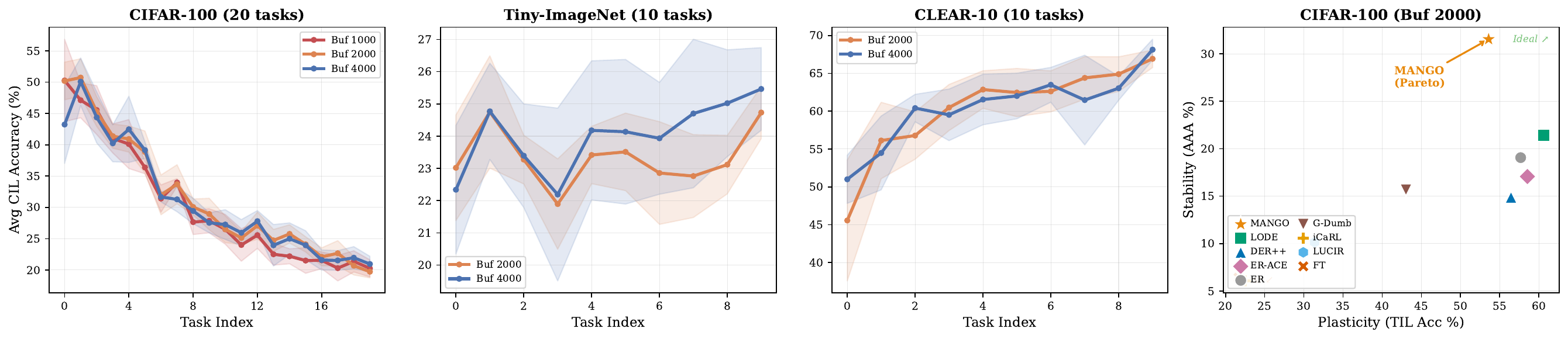}
\caption{From left to right: Task-wise accuracy trajectories for MANGO evaluated on CIFAR-100, Tiny-ImageNet, and CLEAR-10 under varying replay buffer capacities. The minimal variance between the lines indicate MANGO's meta-learned stability mechanism remains robust even in low-memory regimes. The rightmost panel illustrates the stability-plasticity trade-off on CIFAR-100, demonstrating that MANGO optimally balances the retention of past knowledge (y-axis) with the acquisition of new tasks (x-axis) compared to existing state-of-the-art architectures.}
\label{fig:mango_results}
\end{figure}

In this paper, we introduce Meta-Adaptive Network Gradient Optimization (MANGO), a framework for OCL that addresses the limitations by directly controlling parameter  updates in a model based on their impact on past knowledge to prevent CF while learning new tasks efficiently, mitigating the stability-plasticity dilemma as visualized in the right most panel of Figure~\ref{fig:mango_results}.  MANGO introduces an approach which modulates parameter updates and learns new tasks via gradient-gating while dynamically learning layer-wise stability coefficients through meta-learned regularization. Gradient gating selectively scales gradients at parameter level based on normalized parameter magnitude. This allows the model to avoid destructive updates that impact sensitive parameters encoding past knowledge, while allowing adaptability in less critical parameters. Meta-learned regularization dynamically learns layer-wise stability coefficients using a  bi-level meta objective. This directly evaluates the effect of each parameter update on the replay buffer, allowing the model to control the regularization strength based on the evaluated CF rather than fixed heuristics. Essentially, replay acts as both a training signal and a forgetting evaluator, allowing the model to evaluate its own stability rather than operating on fixed heuristics. We evaluate MANGO on three benchmark OCL datasets covering class-incremental learning (CIL) and domain incremental learning (DIL) settings: Split CIFAR-100, Split Tiny-ImageNet, and CLEAR-10 for DIL. MANGO achieves state-of-the-art results across all datasets and buffer sizes, refer Figure~\ref{fig:mango_results} for visualization of accuracies in strict online-settings, surpassing the prior state-of-the-art LODE \citep{liang2023loss} by  5.36\% on CIFAR-100 and 15.18\% on Tiny-ImageNet. Table~\ref{tab:main_results} contains detailed analysis of MANGO's performance across baselines. On CLEAR-10, a practical continual learning benchmark, MANGO strongly achieves positive Backward Transfer (+15.12\%) as reflected in Table~\ref{tab:clear10_results}, indicating that new domain knowledge improves previously learned ones. Positive Backward Transfer is rare in continual learning, reflecting that our meta-learned regularization step enables the model to learn generalizable representations rather than memorizing task patterns. 

We have made the following contributions in our work:
\begin{itemize}
    \item We propose MANGO, a framework for online continual learning that formulates learning as an adaptive and feedback-driven parameter update method.
    \item We introduce a gradient-gating mechanism that modulates parameter updates based on their sensitivity to prevent destructive updates on past tasks.
    \item We developed a meta-learned regularization strategy that adapts stability coefficients using replay-based feedback, evaluating effects of parameter update on previously learned samples.
    \item We demonstrate through extensive experiments and ablation study that MANGO achieves state-of-the-art results across three OCL benchmark datasets, buffers and evaluation metrics, attaining positive Backward Transfer in a DIL online-setting.
\end{itemize}

\section{Related Works}
Existing works in continual learning span broadly across various categories based on their mechanism to mitigate CF.  We distinguish between offline continual learning and OCL. While many approaches are originally formulated for offline settings, we analyze their applicability and limitations in strict online settings. 
Replay-based continual learning approaches maintain an episodic replay memory buffer of past samples and include them in the training process of new tasks. Experience Replay (ER) \citep{chaudhry2019continual, rolnick2019experience, chaudhry2019efficient} utilizes rehearsal by integrating current and replay samples in each mini-batch of incoming training samples. DER++ \citep{buzzega2020dark} is an extension of ER \citep{chaudhry2019continual}, where it uses knowledge distillation by storing and replaying logit outputs, preventing drifts in representation beyond class labels. ER-ACE \citep{caccia2022new} uses asymmetric cross-entropy to separate the current and replay samples for reducing representation errors. GDUMB \citep{prabhu2020gdumb} greedily stores the incoming data in a class-balanced memory buffer and trains solely on the buffer samples. Rehearsal based methods introduce a bias toward stored samples, and revisiting replay samples during training leads to reduced plasticity. 
Projection-based approaches \citep{saha2021gradient, saha2023sgp, farajtabar2020orthogonal} like Gradient Projection Memory (GPM) \citep{saha2021gradient} projects gradient updates onto orthogonal complement of the subspace containing past tasks, safeguarding past tasks. Scaled Gradient Projection (SGP) \citep{saha2023sgp} combines orthogonal projections with gradient steps on sensitive gradient subspaces. While effective in preserving stability, projection-based approaches maintain a growing memory of past subspaces, becoming computationally intensive for strict online settings. 
Regularization based methods \citep{kirkpatrick2017overcoming, aljundi2018memory} penalize parameter drifts of sensitive tasks. Elastic Weight Consolidation (EWC) \citep{kirkpatrick2017overcoming} estimates parameter sensitivity as the diagonal Fisher Information Matrix after each task and penalizes deviation accordingly. LUCIR \citep{hou2019learning} uses cosine normalization and margin based loss. These approaches work on fixed-heuristics which do not adapt during training according to parameter sensitivity across dynamic tasks. Knowledge distillation approach \citep{rebuffi2017icarl, liang2023loss, buzzega2020dark, douillard2020podnet} penalizes changes in output distributions across tasks. iCaRL \citep{rebuffi2017icarl} combines replay with classification and distillation. LODE \citep{liang2023loss} uses loss decoupling by separating current and replay sample contributions. While distillation is effective at preserving prediction stability, they operate at the output distributions, constraining predictions and not parameter updates. Meta-learning methods \citep{finn2017model, saha2025amphibian, riemer2018learning, javed2019meta} adaptively learn tasks, MAML \citep{finn2017model} and its variants present model-agnostic frameworks for fast-adaptation. Amphibian \citep{saha2025amphibian} uses meta-learning to adaptively learn gradient updates in an OCL setting, relying solely on the current data stream with no replay. The absence of replay in \citep{saha2025amphibian} leads to lack of a feedback mechanism to reduce CF over a long sequence of tasks.
Our approach addresses these limitations by controlling updates at parameter level, based on their impact on previously learned tasks to prevent CF and maintain stability while preserving plasticity.

\section{Methodology}
\begin{figure*}[h]
  \centering
  \includegraphics[width=\textwidth]{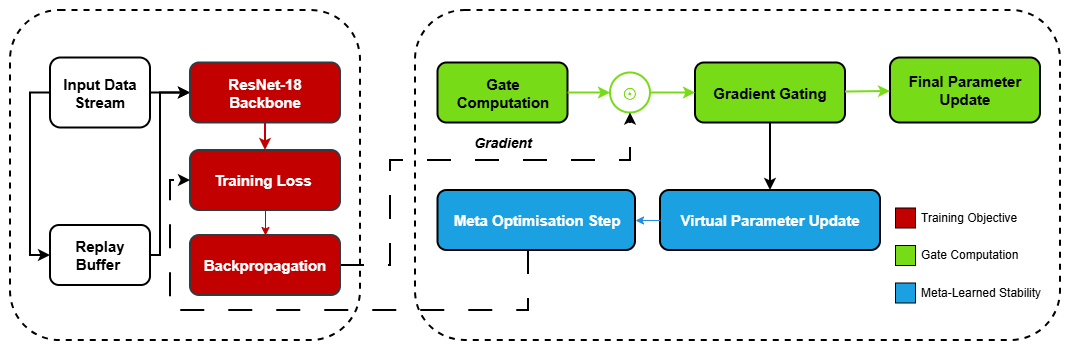}
  \caption{(\textit{Left}) Input stream and replay samples compute the training loss through ResNet-18. (\textit{Right}) Gradients are modulated by per-parameter sigmoid gating to suppress harmful updates. The gated gradient forms a virtual update $\theta'$, evaluated on replay samples to compute $\mathcal{L}_{\text{meta}}$ and adapt the layer-wise stability coefficients $\lambda$ before the final parameter update.}
  \label{fig:architecture}
\end{figure*}
In this section we present the details of MANGO, an OCL framework that addresses the stability-plasticity dilemma through  \emph{gradient gating} and \emph{meta-learned regularization}.  
Our key idea is to control parameter updates at two levels: (i) Gradient-Gating: modulating gradients  at parameter level to prevent harmful updates, and (ii) Meta-learned Regularization: adapting regularization coefficients based on their impact on past knowledge.
\subsection{Problem Setup}

Online continual learning models process a sequence of tasks ${\mathcal{D}_1, \dots, \mathcal{D}_T}$ in an incremental class or domain setting, where data of each task is seen only once. The model receives a mini-batch $(x_t, y_t)$ from the current task at each step $t$, and maintains a replay memory buffer  $\mathcal{M}$  containing samples from past tasks.  
The objective of our work is to learn a model $f\theta$ that performs well on all past tasks without access to full historical data. The buffer $\mathcal{M}$ is maintained via reservoir sampling~\citep{vitter1985random}; during the first mini-batches of Task 1 when $\mathcal{M}$ is empty, the meta-objective is 
skipped and only the gated update is applied.

\subsection{Training Objective}

The proposed method begins with a cross-entropy loss along with a regularization term inspired by Elastic Weight Consolidation (EWC) \citep{kirkpatrick2017overcoming}, that balances learning new tasks while preserving past ones:
\vspace{-1mm}
\begin{equation}
\mathcal{L}_{train} =
\mathcal{L}_{CE}(x_t, y_t)
+ \sum_i \frac{\lambda_i}{2} ||\theta_i - \theta_i^{old}||^2
  \end{equation}

where $\theta^{old}$ represents the previously learned parameters, and $\lambda_i$ are the layer-wise regularization coefficients, which control the stability of the parameters. 
Larger $\lambda_i$ enforces stronger retention of past parameter 
groups; smaller values allow plasticity. Unlike EWC~\citep{kirkpatrick2017overcoming}, 
where $\lambda_i$ is a fixed constant, MANGO learns $\lambda_i$ 
dynamically via the meta-objective described in later sections.

\subsection{Gradient Gating}
Gradient-gating addresses a key limitation of standard training where all the parameters are updated uniformly, which lead to CF. We recognize that some parameters are more critical for stability and should be updated according to their sensitivity to past tasks to prevent harmful updates.
\paragraph{Defining Harmful Updates}
A parameter update $\Delta\theta_j = -\eta g_j$ is considered harmful if it increases the loss on previously learned tasks:
\begin{equation}
\nabla_{\theta_j}\mathcal{L}_{\text{past}}
\cdot
\Delta\theta_j > 0,
\label{eq:harmful}
\end{equation}
where $\mathcal{L}_{\text{past}}$ denotes replay loss on previously learned data and  $\theta_j$ denotes an individual parameter.
Without any gating, a large gradient update may affect parameters important to past tasks, increasing replay loss and leading to CF.

\paragraph{Gradient Gating}
As a solution to this limitation, gradient-gating selectively modulates parameter updates, scaling gradients based on parameter sensitivity.  Let $g = \nabla\theta \mathcal{L}_{train}$ denote the gradient of the training objective. We compute a gated gradient as $\tilde{g}$: 

\begin{equation}
\tilde{g}_j  = g_j
\cdot
\sigma\!\left(
\frac{\theta_j}
{\mathrm{std}(\boldsymbol{\theta}_l)}
\right),
\label{eq:gate_param}
\end{equation}

where $\sigma(\cdot)$ is the sigmoid function and $\boldsymbol{\theta}_l=\{\theta_1,\dots,\theta_n\}$ denotes all parameters in layer $l$. The effective individual parameter update becomes:
\begin{equation}
\tilde{\Delta}\theta_j = -\eta \tilde{g}_j.
\end{equation}

As $\sigma(\cdot)\in(0,1)$,  gradient gating scales updates according to the normalized parameter magnitude within each layer. Parameters receiving smaller gate values receive small updates, limiting destructive parameter drift on sensitive parameters. This lowers the probability of satisfying Eq.~\eqref{eq:harmful}.
\begin{algorithm}[h]
\caption{MANGO}
\label{alg:mango}
\begin{algorithmic}[1]

\Require Model parameters $\theta$, stability coefficients $\lambda$, replay buffer $\mathcal{M}$
\Require Learning rates $\eta, \eta_\lambda$

\For{each task $t = 1, \dots, T$}
    \For{each mini-batch $(x_t, y_t)$}

        \State \textbf{Compute training loss:}
        \State $\mathcal{L}_{train} = \mathcal{L}_{CE}(x_t, y_t) + \sum_i \frac{\lambda_i}{2} \|\theta_i - \theta_i^{old}\|^2$

        \State \textbf{Compute gradients:}
        \State $g \leftarrow \nabla_\theta \mathcal{L}_{train}$

        \State \textbf{Gradient gating:}
        \State $\tilde{g} \leftarrow g \odot \sigma\left(\frac{\theta}{\mathrm{std}({\theta}_l)}\right)$
        \State \textbf{Virtual update (for meta-learning):}
        \State $\theta' \leftarrow \theta - \eta \tilde{g}$

        \State Sample replay batch $(x_{mem}, y_{mem}) \sim \mathcal{M}$

        \State \textbf{Meta-objective:}
        \State $\mathcal{L}_{meta} \leftarrow \mathcal{L}_{CE}(x_{mem}, y_{mem}; \theta')$

        \State \textbf{Update stability coefficients:}
        \State $\lambda \leftarrow \lambda - \eta_\lambda \nabla_\lambda \mathcal{L}_{meta}$

        \State \textbf{Final parameter update:}
        \State $\theta \leftarrow \theta - \eta \tilde{g}$

        \State Update replay buffer $\mathcal{M}$ (reservoir sampling)

    \EndFor
\EndFor

\end{algorithmic}
\end{algorithm}
\subsection{Meta-Learned Stability}
\label{meta}
While gradient-gating controls parameter updates, it needs a mechanism to determine how strongly different parameter groups should be protected. Fixed stability coefficients are insufficient in adapting varying parameter sensitivity across tasks. To address this, we introduce meta-learned regularization, which adapts the stability coefficients  $\lambda_i$ using a replay-based meta-objective to directly measure CF.

At each step, we create a \emph{virtual parameter} ($\theta'$) using gated gradient as the training step: \begin{equation} \theta' = \theta - \eta \tilde{g} \end{equation}
We then evaluate the effect of this virtual update on replay samples $(x_{mem}, y_{mem})$ from the buffer $\mathcal{M}$ :
\begin{equation}
\mathcal{L}_{meta} =
\mathcal{L}_{CE}(x_{mem}, y_{mem}; \theta')
\end{equation}
The regularization coefficients $\lambda$ are updated through the meta-objective with learning rate  $\eta_\lambda$:
\begin{equation}
\lambda \leftarrow \lambda - \eta_\lambda \nabla_\lambda \mathcal{L}_{meta}
\end{equation}
As the virtual update ($\theta'$) depends on the stability coefficients ($\lambda$) through $\mathcal{L}_{train}$, the gradients from the meta-objective propagate back to $\lambda$. This enables the model to increase stability for sensitive parameters while allowing less sensitive parameters to remain flexible, reducing CF.
\paragraph{Why this prevents forgetting}

The training objective contains the regularization term which penalizes drift from past parameters:
\begin{equation}
\sum_i
\frac{\lambda_i}{2}
\|
\theta_i-\theta_i^{\text{old}}
\|^2,
\end{equation}

The meta-gradient can be written as:
\begin{equation}
\nabla_\lambda \mathcal{L}_{\text{meta}}
=
\frac{\partial \mathcal{L}_{\text{meta}}}{\partial \theta'}
\frac{\partial \theta'}{\partial \lambda}.
\label{eq:meta_chain}
\end{equation}
\vspace{0.1mm}
\subsection{Discussion} When replay loss increases after a virtual update, the corresponding $\lambda_i$ increases, strengthening regularization and reducing future drift from $\theta_i^{\text{old}}$. Conversely, when replay loss is stable, $\lambda_i$ decreases, allowing greater plasticity. The layer-wise evolution of the meta-learned $\lambda_i$ and a heatmap confirming that MANGO successfully retains knowledge of older tasks throughout the training is reported in Figure~\ref{fig:lambda.png}.
Creating an adaptive stability--plasticity mechanism where replay feedback dynamically controls regularization strength instead of relying on fixed heuristics.

\section{Experiments}

\subsection{Experimental Setup}

\textbf{Datasets:} We performed experimentation over three standard benchmark datasets for OCL. For CIL, we use Split CIFAR-100 \citep{krizhevsky2009learning} and Split Tiny-ImageNet \citep{le2015tiny}. For DIL, we use CLEAR-10 \citep{lin2021clear}, a dataset specifically designed for continual learning evaluation. CIFAR-100 is divided into 20 tasks with 5 classes per task, Tiny-ImageNet into 10 tasks with 20 classes per task, and CLEAR-10 into 10 domains.

\textbf{Model and training:} All experiments use ResNet-18~\citep{he2016deep}. For CIFAR-100, we use a $3\times3$ kernel and remove max pooling; Tiny-ImageNet and CLEAR-10 use the standard architecture. Training follows the strict online setting with a single epoch per task and 3 glances per mini-batch \citep{caccia2020online}. We use SGD with momentum 0.9, no weight decay, online and replay batch sizes are 32 and 64 respectively, and replay buffers of 2000 and 4000 samples. MANGO combines gradient gating with meta-learned regularization to reduce catastrophic forgetting. All experiments are implemented in PyTorch (CUDA-enabled) and averaged over 5 random seeds. 

\textbf{Baselines:} We compare against replay, distillation, and regularization-based continual learning methods, including ER and its variants \citep{buzzega2020dark,caccia2022new}, LODE~\citep{liang2023loss}, LUCIR~\citep{hou2019learning}, iCaRL~\citep{rebuffi2017icarl}, GDUMB~\citep{prabhu2020gdumb}, and Fine-Tuning (FT) as a no-replay lower bound.

\textbf{Metrics and Evaluation:} We report Accuracy (Acc), Average Anytime Accuracy (AAA)~\citep{caccia2020online}, Worst-case Accuracy (WC-Acc)~\citep{delange2023continual}, and Backward Transfer (BWT). Following prior OCL work, Acc, AAA, and WC-Acc are the primary metrics for strict online CIL, while BWT is reported in the appendix for completeness. Additional analyses, including small-buffer experiments, BWT and TIL evaluation, and experimental setup are also provided in the Appendices~\ref{app:additional}, \ref{app:bwt}, and \ref{app:impl}.
\begin{figure}[b]
\centering
\includegraphics[width=0.95\columnwidth]{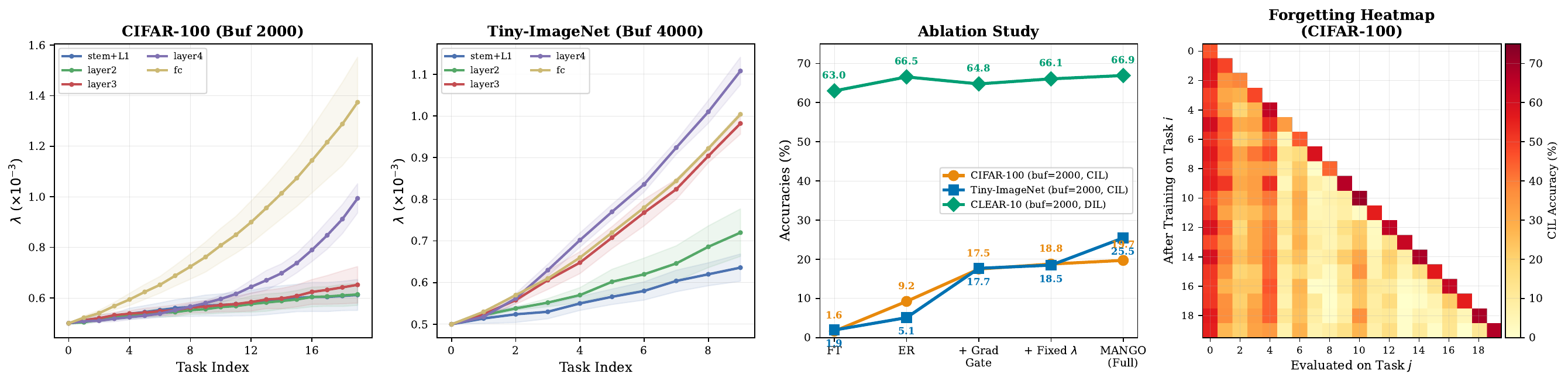}
\caption{From left to right: The first two panels plot the dynamic, layer-wise evolution of the meta-learned $\lambda$, as new tasks are introduced. The third panel demonstrates the progressive accuracy improvements yielded by our proposed gating mechanism and adaptive regularization compared to baseline fine-tuning and replay (ER). The final panel shows a CIL accuracy heatmap on CIFAR-100, confirming that MANGO successfully retains knowledge of older tasks throughout the training.}
\label{fig:lambda.png}
\end{figure}

\begin{table*}[t]
\caption{%
  Comparison on Split CIFAR-100 and Split Tiny-ImageNet,  single-pass results over 5 seeds.\\}
\label{tab:main_results}
\centering
\setlength{\tabcolsep}{4pt}
\renewcommand{\arraystretch}{1.06}

\begin{tabular*}{\textwidth}{@{\extracolsep{\fill}}
  l l
  r r r
  r r r
  @{}}
\toprule
& & \multicolumn{3}{c}{\textbf{CIFAR-100} (CIL)}
  & \multicolumn{3}{c}{\textbf{TinyImageNet} (CIL)} \\
\cmidrule(lr){3-5}\cmidrule(lr){6-8}
\multirow{2}{*}{Method}
  & \multirow{2}{*}{Buf.}
  & {Acc} & {AAA } & {WC-Acc }
  & {Acc} & {AAA } & {WC-Acc } \\
  & & {\small(\%)} & {\small(\%)} & {\small(\%)}
    & {\small(\%)} & {\small(\%)} & {\small(\%)} \\
\midrule

\multicolumn{8}{l}{\textit{Buffer\,=\,2\,000}} \\[1pt]
FT      & ---
  & $1.65${\tiny$\pm.29$}  & $6.41${\tiny$\pm.46$}  & 2.11{\tiny$\pm$}
  & $1.94${\tiny$\pm.40$} &$4.86${\tiny$\pm.43$} &$2.11${\tiny$\pm.28$} \\
ER \citep{chaudhry2019continual}    & 2k
  & $9.25${\tiny$\pm.63$}  & $19.06${\tiny$\pm.65$} & $9.00${\tiny$\pm.77$}
  & 5.10{\tiny$\pm.15$} & $4.86${\tiny$\pm.33$} &$2.11${\tiny$\pm.20$} \\
ER-ACE \citep{caccia2022new} & 2k 
  & $11.77${\tiny$\pm1.56$} & $17.07${\tiny$\pm2.09$} & $7.28${\tiny$\pm1.22$}
  & $5.10${\tiny$\pm.14$}& $10.75${\tiny$\pm.32$} & $4.83${\tiny$\pm.21$} \\
GDUMB \citep{prabhu2020gdumb} & 2k
  & $5.35${\tiny$\pm.64$}  & $15.66${\tiny$\pm.28$} & $8.05${\tiny$\pm.60$}
  & $1.58${\tiny$\pm.23$}  & $5.68${\tiny$\pm.29$} & $4.14${\tiny$\pm0.23$} \\
iCaRL \citep{rebuffi2017icarl}  & 2k
  & $1.39${\tiny$\pm.02$}  & $6.07${\tiny$\pm.23$}  & $3.19${\tiny$\pm.08$}
  & $.61${\tiny$\pm.06$}  & $1.86${\tiny$\pm.07$} & $1.42${\tiny$\pm0.11$} \\
LUCIR \citep{hou2019learning}  & 2k
  & $2.88${\tiny$\pm.44$}  & $10.14${\tiny$\pm.52$} & $4.00${\tiny$\pm.29$}
  & $3.05${\tiny$\pm.26$}  & $8.14${\tiny$\pm.28$} & $4.07${\tiny$\pm0.30$} \\
DER++ \citep{buzzega2020dark}   & 2k
  & $6.54${\tiny$\pm1.34$} & $14.89${\tiny$\pm1.10$} & $5.00${\tiny$\pm.58$}
  & $6.62${\tiny$\pm.99$}  & $12.05${\tiny$\pm.63$} & $6.73${\tiny$\pm0.49$} \\
LODE \citep{liang2023loss}   & 2k
  & $14.36${\tiny$\pm.70$} & $21.42${\tiny$\pm1.33$} & $12.36${\tiny$\pm1.86$}
  & $9.97${\tiny$\pm.30$}  & $13.96${\tiny$\pm.83$} & $9.12${\tiny$\pm0.52$} \\
MANGO   & 2k
  & $\mathbf{19.72}${\tiny$\pm1.01$} & $\mathbf{31.53}${\tiny$\pm.37$} & $\mathbf{12.69}${\tiny$\pm.55$}
  & $\mathbf{25.15}${\tiny$\pm1.32$}  & $\mathbf{23.45}${\tiny$\pm1.13$} & $\mathbf{17.44}${\tiny$\pm0.84$} \\

\midrule

\multicolumn{8}{l}{\textit{Buffer\,=\,4\,000}} \\[1pt]
ER   \citep{chaudhry2019continual}   & 4k
  & $12.32${\tiny$\pm.42$} & $19.30${\tiny$\pm1.29$} & $8.00${\tiny$\pm.90$}
  & 5.93{\tiny$\pm.34$} & $12.02${\tiny$\pm.15$} &$6.62${\tiny$\pm.17$}\\
ER-ACE \citep{caccia2022new} & 4k
  & $12.24${\tiny$\pm2.28$}& $17.99${\tiny$\pm.93$} & $7.74${\tiny$\pm1.19$}
  & $5.93${\tiny$\pm.35$}& $12.02${\tiny$\pm.15$} & $6.62${\tiny$\pm.17$} \\
GDUMB \citep{prabhu2020gdumb} & 4k
  & $9.43${\tiny$\pm.74$}  & $20.18${\tiny$\pm.80$} & $11.02${\tiny$\pm0.57$}
  & $2.68${\tiny$\pm.20$}  & $8.23${\tiny$\pm.21$} & $6.09${\tiny$\pm0.47$}\\
iCaRL \citep{rebuffi2017icarl}  & 4k
  & $1.67${\tiny$\pm.23$}  & $6.73${\tiny$\pm.34$}  & $3.13${\tiny$\pm.48$}
  &$.64${\tiny$\pm.09$}  & $2.30${\tiny$\pm.12$} & $1.68${\tiny$\pm0.16$} \\
LUCIR  \citep{hou2019learning} & 4k
  & $2.88${\tiny$\pm.30$}  & $9.81${\tiny$\pm.74$}  & $4.33${\tiny$\pm.17$}
  & $2.84${\tiny$\pm.27$}  & $8.27${\tiny$\pm.18$} & $4.19${\tiny$\pm0.18$} \\
DER++ \citep{buzzega2020dark}  & 4k
  & $5.08${\tiny$\pm.76$}  & $14.76${\tiny$\pm.43$} & $4.63${\tiny$\pm.26$}
  & $6.90${\tiny$\pm.53$}  & $12.95${\tiny$\pm.54$} & $7.47${\tiny$\pm0.55$} \\
LODE \citep{liang2023loss}   & 4k
  & $15.82${\tiny$\pm1.28$} & $21.44${\tiny$\pm3.02$} & $11.31${\tiny$\pm3.24$}
  & $11.68${\tiny$\pm.37$}  & $15.74${\tiny$\pm.97$} & $11.64${\tiny$\pm0.93$} \\
MANGO   & 4k
  & $\mathbf{20.64}${\tiny$\pm1.15$} & $\mathbf{30.67}${\tiny$\pm.95$} & $\mathbf{11.60}${\tiny$\pm1.40$}
  & $\mathbf{24.73}${\tiny$\pm.80$}  & $\mathbf{23.33}${\tiny$\pm.37$} & $\mathbf{17.27}${\tiny$\pm0.68$} \\

\bottomrule
\end{tabular*}
\end{table*}

\subsection{Results on CIFAR-100}
Table \ref{tab:main_results} displays the baseline results on CIFAR-100, on 20 tasks on CIL. Our method substantially outperforms all baselines across standard OCL evaluation metrics, presenting robustness against replay buffer size. On a replay memory buffer of 2000, MANGO achieves an accuracy of 19.72 ± 1.01\%, representing a  37.3\% relative error reduction compared to the previous state-of-the-art method LODE \citep{liang2023loss} (14.36 ± 0.70\%). This method’s superiority is further reflected in AAA, where MANGO attains 31.53 ± 0.37\% compared to LODE’s \citep{liang2023loss} 21.42 ± 1.33\%, a hike of over 10 percentage indicating that our method’s performance is superior not just in final accuracy but is consistently maintained across all tasks. We achieve 12.69 ± 0.55\% WC-Acc, displaying the highest worst-case accuracy across all tasks, penalizing CF most severely. Figure~\ref{fig:mango_results} shows MANGO’s strong task-wise performance across both buffer sizes. proving MANGO’s robustness to memory constraints Which is a critical property for OCL deployment. The performance of our method reflects the importance of the combination of gradient gating and meta-learned regularization, as it allows the model to reduce CF without compromising plasticity.
\subsection{Results on Tiny-ImageNet}
Tiny-ImageNet is a significantly difficult benchmark, being split across 10 tasks over 200 classes and high similarity between classes. Table \ref{tab:main_results} demonstrates that MANGO delivers the highest performance gains on this dataset compared to its baseline methods. With replay memory buffer 4000, MANGO achieves an accuracy of 24.73 ± 0.80\%, a 13.05\% absolute error reduction compared to its best competing method LODE \citep{liang2023loss}. This dramatic gap between the performance of the two methods confirms that the adaptive nature of MANGO’s gradient-gating and meta-learned regularization performs exceptionally well on harder, more detailed continual learning settings where fixed heuristics in existing methods fall behind. Results reflect that MANGO leads on AAA (23.33 ± 0.37\%) and WC-Acc (17.27 ± 0.68\%) as well by large margins, demonstrating more stable learning across all tasks and  effectiveness at preventing collapse of performance on any single past task. The results on Tiny-ImageNet verify the claim that meta-learned and feedback based regularization outperforms the existing methods and stand superior in addressing CF in a strict online setting. Figure~\ref{fig:mango_results} shows the robust performance of Tiny-ImageNet across tasks and buffers.

\begin{table*}[t]
\caption{%
  Comparison on CLEAR-10, single-pass  results over 5 seeds.\\}
  \vspace{2mm}
\label{tab:clear10_results}
\centering
\setlength{\tabcolsep}{4pt}
\renewcommand{\arraystretch}{1.06}
 
\begin{tabular}{lccccc}
\toprule
& & \multicolumn{4}{c}{\textbf{CLEAR-10 (DIL)}} \\
\cmidrule(lr){3-6}
\multirow{2}{*}{Method}
  & \multirow{2}{*}{Buf.}
  & {Acc } & {AAA } & {WC-Acc } & {BWT } \\
  & & {\small(\%)} & {\small(\%)} & {\small(\%)} & {\small(\%)} \\
\midrule
 
\multicolumn{6}{l}{\textit{Buffer\,=\,2\,000}} \\[1pt]
FT      & ---
  & $62.99${\tiny$\pm.64$}  & $61.78${\tiny$\pm.91$}  & $53.02${\tiny$\pm1.87$}  & ${-}$ \\
ER   \citep{chaudhry2019continual}   & 2k
  & $65.21${\tiny$\pm1.02$}  & $61.89${\tiny$\pm1.78$}  & $\mathbf{60.64}${\tiny$\pm1.43$}  & $-10.00${\tiny$\pm3.75$} \\
ER-ACE \citep{caccia2022new}  & 2k
  & $58.38${\tiny$\pm.60$}  & $57.10${\tiny$\pm.92$}  & $48.78${\tiny$\pm1.52$}  & ${-}$ \\
GDUMB \citep{prabhu2020gdumb} & 2k
  & $10.41${\tiny$\pm.27$}  & $10.40${\tiny$\pm.25$}  & $9.82${\tiny$\pm.25$}    & ${-}$ \\
DER++ \citep{buzzega2020dark}  & 2k
  & $59.89${\tiny$\pm.80$}  & $58.55${\tiny$\pm.98$}  & $51.52${\tiny$\pm1.92$}  & ${-}$ \\
LODE  \citep{liang2023loss}  & 2k
  & $64.23${\tiny$\pm1.37$} & $64.18${\tiny$\pm1.33$} & $51.26${\tiny$\pm1.60$} & $0.52${\tiny$\pm6.5$} \\
MANGO   & 2k
  & $\mathbf{66.91}${\tiny$\pm1.20$} & $\mathbf{66.35}${\tiny$\pm1.05$} & $55.0${\tiny$\pm1.1$} & $\mathbf{+15.12}${\tiny$\pm3.3$} \\
 
\midrule
 
\multicolumn{6}{l}{\textit{Buffer\,=\,4\,000}} \\[1pt]
ER \citep{chaudhry2019continual}     & 4k
  & $67.10${\tiny$\pm1.95$} & $59.26${\tiny$\pm2.38$} & $\mathbf{61.34}${\tiny$\pm2.36$} & $-9.67${\tiny$\pm4.88$} \\
ER-ACE \citep{caccia2022new} & 4k
  & $58.10${\tiny$\pm1.23$} & $56.5${\tiny$\pm1.2$}   & $50.0${\tiny$\pm2.0$}    & ${-}$ \\
GDUMB \citep{prabhu2020gdumb} & 4k
  & $10.41${\tiny$\pm.27$}  & $10.40${\tiny$\pm.25$}  & $9.82${\tiny$\pm.25$}    & ${-}$ \\
DER++ \citep{buzzega2020dark}  & 4k
  & $60.13${\tiny$\pm.75$}  & $58.5${\tiny$\pm1.0$}   & $53.0${\tiny$\pm1.5$}    & ${-}$ \\
LODE \citep{liang2023loss}   & 4k
  & $64.34${\tiny$\pm2.43$} & $64.34${\tiny$\pm2.43$} & $51.08${\tiny$\pm1.89$} & $3.96${\tiny$\pm8.5$} \\
MANGO   & 4k
  & $\mathbf{67.67}${\tiny$\pm1.94$} & $\mathbf{67.05}${\tiny$\pm1.6$} & $55.3${\tiny$\pm2.5$} & $\mathbf{+13.5}${\tiny$\pm5.8$} \\
 
\bottomrule
\end{tabular}
\end{table*}

\subsection{Results on CLEAR-10}
CLEAR-10 is a DIL benchmark designed to simulate realistic distribution shifts over time, making it a more practical evaluation than split benchmarks. CLEAR-10 has a fixed class space while its input distribution changes across 10 tasks. Figure~\ref{fig:mango_results} demonstrates the consistently rising performance of across tasks. Table \ref{tab:clear10_results} show that with a replay buffer of 2000, MANGO attains highest accuracy of 66.91 ± 1.20\%, outperforming the previous state-of-the-art method LODE \citep{liang2023loss} with a gain of ~2\%.  Fine-Tuning (FT) does surprisingly well in DIL settings due to the shared class spaces across tasks. MANGO also achieves strong positive backward transfer (+15.12 ± 3.3\%), indicating that learning new domains improves performance on previous ones. Achieving a positive BWT is rare in continual learning, reflecting that our meta-learned regularization step enables the model to learn generalizable representations rather than memorizing task patterns. In contrast, competing methods such as DER++~\citep{buzzega2020dark}, ER-ACE~\citep{caccia2022new}, and LODE achieve near-zero BWT, suggesting limited forward transfer across domains. MANGO attains the highest AAA as well, demonstrating stable performance throughout training. With a replay buffer of 4000, MANGO achieves the highest Acc, AAA, and positive BWT, showing consistent performance across both small and large replay settings. These results demonstrate that MANGO generalizes effectively across both CIL and DIL continual learning benchmarks.

\subsection{Ablation Study and Analysis}
\label{sec:ablation}
We perform an ablation study across all three benchmarks and replay buffers to understand the individual contribution of each component of MANGO. In our study, we evaluate four components of MANGO: (i) w/o Replay (FT): Fine-Tuning with no replay, this acts as a lower-bound for CF. (ii) w/o Meta-Lambda: evaluating MANGO without meta-learned regularization, using gradient gating along without adaptive stability coefficients. (iii) w. Only Replay (ER): evaluating MANGO on standard ER \citep{chaudhry2019continual} without gradient gating or meta-learned regularization. (iv) w/o Regularization: MANGO with gradient gating without any regularization term. The results of the ablation study on the buffer size 2000 are demonstrated in Table \ref{tab:ablation}, Figure~\ref{fig:lambda.png}, and additional buffer analysis is in the Appendix~\ref{app:buffer}.

\textbf{Impact of Meta-Learned Regularization:} Removing meta-lambda reduces CIL Acc on CIFAR-100 and Tiny-ImageNet from 19.72\% to 18.79\% and 25.46\% to 18.50\% respectively. The larger drop on harder dataset reflects that adaptive stability coefficients are critical when learning more challenging datasets, where fixed heuristics are insufficient. On CLEAR-10, the performance drop is smaller, likely due to the shared class space in DIL settings.  The consistent performance drop across all datasets confirms that meta-learning the regularization coefficients is a generalizable strength of MANGO’s performance and is not dataset-specific. 

\textbf{Effect of gradient-gating:} We compare MANGO with w/o Regularization (gating only) and Replay-only ER. This comparison helps us evaluate the impact of gradient-gating. Using gradient gating without regularization reduces performance on CIFAR-100 and Tiny-ImageNet, showing that gating alone improves over replay but requires adaptive regularization for best performance. The progression from Replay-only ER, to gating-only, to w/o Meta-Lambda, and finally MANGO highlights the contribution of each component.

\textbf{Role of Replay:} Comparing MANGO with Fine-Tuning (FT) demonstrates the importance of replay as a feedback signal.  Without replay, CIFAR-100 and Tiny-ImageNet accuracies drop sharply to 1.65\% and  1.94\% respectively. However, replay alone is also insufficient, as MANGO further uses replay samples within the meta-objective to directly evaluate and minimize forgetting during training. The consistent improvements across both CIL and DIL benchmarks demonstrate the robustness of the proposed framework.
\begin{table*}[t]
\caption{%
  Ablation results on 2000 buffer CIFAR-100, Tiny-ImageNet, and CLEAR-10 over 5 seeds.}
  \vspace{2mm}
\label{tab:ablation}
\centering
\setlength{\tabcolsep}{4pt}
\renewcommand{\arraystretch}{1.06}
 
\begin{tabular}{lcccc}
\toprule
& & \multicolumn{1}{c}{\textbf{CIFAR-100}}
  & \multicolumn{1}{c}{\textbf{TinyImageNet}}
  & \multicolumn{1}{c}{\textbf{CLEAR-10}} \\
\cmidrule(lr){3-3}\cmidrule(lr){4-4}\cmidrule(lr){5-5}
\multirow{2}{*}{Method}
  & \multirow{2}{*}{Buf.}
  & {CIL Acc}
  & {CIL Acc}
  & {DIL Acc} \\
  & & {\small(\%)} & {\small(\%)} & {\small(\%)} \\
\midrule
 

MANGO               & 2k
  & $\mathbf{19.72}${\tiny$\pm1.01$}    & $\mathbf{25.46}${\tiny$\pm1.36$}   & $\mathbf{66.91}${\tiny$\pm1.20$} \\
w/o Replay (FT)     & ---
  & $1.65${\tiny$\pm.29$}               & $1.94${\tiny$\pm.40$}              & $62.99${\tiny$\pm.64$} \\
w/o Meta-$\lambda$  & 2k
  & $18.79${\tiny$\pm2.34$} & $18.50${\tiny$\pm.66$}             & $66.06${\tiny$\pm3.56$} \\
w/ only Replay (ER) & 2k
  & $9.25${\tiny$\pm.63$}               & $5.10${\tiny$\pm.15$}              & $66.55${\tiny$\pm2.35$} \\
w/o Regularization  & 2k
  & $17.51${\tiny$\pm2.14$}             & $17.70${\tiny$\pm.70$} & $64.77${\tiny$\pm3.11$} \\
 
\bottomrule
\end{tabular}
\end{table*}

\section{Conclusion}
In this work, we proposed MANGO, a framework for online continual learning to address the stability-plasticity dilemma in class and domain incremental learning. We modulate  parameter updates at each step based on normalized sensitivity, protecting plasticity. We also introduce meta-learned regularization where we adapt layer-wise stability coefficients that directly measure CF, preserving stability. Experiments reflect that our method achieves state-of-the-art results, outperforming all strong baselines across three standard benchmarks and multiple replay sizes. Notably, MANGO achieves strong positive Backward Transfer on domain incremental tasks.

\section{Limitations}
Our current setup assumes a well defined task and domain boundaries, which may not uphold in task-agnostic settings. In our future works we will extend our method to a task-free and blurry-boundary continual learning setup and pre-train foundation models. We hope MANGO can contribute to building one-shot machine learning systems that can learn continually from real-world data.

\bibliographystyle{unsrtnat}
\bibliography{references}

\appendix

\section*{Appendix}
\addcontentsline{toc}{section}{Appendix}
We are providing supplementary material and additional experimentation information in this appendix.
The appendix is divided into four sections.
\textbf{Appendix~\ref{app:additional}} provides additional quantitative results including additional buffer analysis, task-incremental learning (TIL) accuracies over datasets along with a dataset summary.
provides additional quantitative results
\textbf{Appendix~\ref{app:bwt}} reports the Backward Transfer (BWT) analysis for class-incremental learning (CIL) over CIFAR-100 and Tiny-ImageNet along with discussion over its limited informativeness in an online setting.
\textbf{Appendix~\ref{app:impl}} includes all details regarding experimentation necessary for reproducing our method, including hyperparameters, architecture configuration and replay memory buffer management.
\textbf{Appendix~\ref{app:ablation}} contains an extended ablation study over additional replay memory buffer.

\section{Additional Quantitative Results}
\label{app:additional}

\subsection{Small Buffer (M = 1000) Results}
\label{app:small_buffer}

Table~\ref{tab:small_buffer} presents CIL accuracy results at buffer
size $M{=}1000$ across CIFAR-100, extending the main paper's
$M{=}2000$ and $M{=}4000$ evaluations to a more severely replay constrained
online setting. 

MANGO achieves $20.25 \pm 1.38\%$ on CIFAR-100 at $M{=}1000$,  surpassing the accuracies of other baseline mathods, confirming that gradient gating along with meta-learned regularization compensates for reduced replay memory by modulating the stability parameters based on their sensitivity.

\begin{table}[h]
\caption{Results on CIFAR-100 at small buffer $M{=}1000$, single-pass over 5 seeds.}
\label{tab:small_buffer}
\centering
\small
\begin{tabular}{lc}
\toprule
Method & \textbf{CIFAR-100} CIL (\%) \\
\midrule
FT (no replay)   &$1.65_{\pm 0.29}$\\
ER  \citep{chaudhry2019continual}             &$6.55_{\pm 0.41}$\\
ER-ACE \citep{caccia2022new}          &$11.30_{\pm 0.85}$\\
GDUMB \citep{prabhu2020gdumb}          &$4.57_{\pm 0.56}$\\
iCaRL \citep{rebuffi2017icarl}          &$1.22_{\pm 0.08}$\\
LUCIR  \citep{hou2019learning}           &$2.77_{\pm 0.27}$\\
DER++ \citep{buzzega2020dark}           &$6.93_{\pm 0.53}$\\
LODE  \citep{liang2023loss}            &$13.30_{\pm  0.55}$\\
\textbf{MANGO (ours)}    & $\mathbf{20.25}_{\pm 1.38}$ \\
\bottomrule
\end{tabular}
\end{table}

\subsection{Task-Incremental Learning (TIL) Evaluation}
\label{app:til}

Task-incremental learning (TIL) is a continual learning setting where the identity of the task is available to the model during testing. This allows the model to use the accurate classification labels. TIL accuracy measures representation quality among tasks, avoiding harder problems like cross-task class separation.  Table~\ref{tab:til} evaluates MANGO along with all the baselines on TIL across CIFAR-100 and Tiny-ImageNet at buffer sizes $M{=}2000$ and $4000$. 
Interestingly, the results on TIL reflect a notable difference in performance between MANGO and LODE \citep{liang2023loss} on CIFAR-100, LODE \citep{liang2023loss} surpasses MANGO despite having lower CIL accuracy. This shows that LODE' \citep{liang2023loss} loss-decoupling approach efficiently preserves task representations, resulting in stronger distinguishment within tasks but poor inter-task separation.  However, MANGO surpasses LODE \citep{liang2023loss} on Tiny-ImageNet at Buffer 2000, confirming that this trade-off depends on difficulty of datasets.

\begin{table}[h]
\caption{Task-incremental learning (TIL) results, single-pass over 5 seeds.}
\label{tab:til}
\centering
\small
\setlength{\tabcolsep}{4pt}
\begin{tabular}{lccc}
\toprule
Method & Buf. & \textbf{CIFAR-100} (\%) & \textbf{Tiny-ImageNet} (\%) \\
\midrule
\multicolumn{4}{l}{\textit{Buffer $= 2{,}000$}} \\[2pt]
FT    & ---  & $21.67_{\pm 1.31}$ & $7.82_{\pm 0.95}$  \\
ER \citep{chaudhry2019continual}   & 2k   & $57.68_{\pm 0.68}$ & $27.26_{\pm 0.32}$ \\
ER-ACE \citep{caccia2022new} & 2k  & $58.54_{\pm 2.32}$ & $27.26_{\pm 0.33}$ \\
GDUMB \citep{prabhu2020gdumb} & 2k  & $43.04_{\pm 2.14}$ & $10.51_{\pm 0.51}$ \\
iCaRL \citep{rebuffi2017icarl} & 2k  & $22.50_{\pm 0.05}$ & $5.70_{\pm 0.43}$  \\
LUCIR \citep{hou2019learning} & 2k  & $31.37_{\pm 1.80}$ & $19.74_{\pm 0.86}$ \\
DER++ \citep{buzzega2020dark} & 2k  & $56.45_{\pm 1.46}$ & $28.38_{\pm 1.60}$ \\
LODE \citep{liang2023loss}  & 2k  & $\mathbf{60.65}_{\pm 1.02}$ & $30.65_{\pm 0.61}$ \\
MANGO (ours) & 2k & $53.59_{\pm 2.38}$ & $\mathbf{33.24}_{\pm 1.27}$ \\
\midrule
\multicolumn{4}{l}{\textit{Buffer $= 4{,}000$}} \\[2pt]
ER \citep{chaudhry2019continual} & 4k   & $63.02_{\pm 0.67}$ & $28.35_{\pm 0.51}$ \\
ER-ACE \citep{caccia2022new} & 4k  & $60.42_{\pm 3.45}$ & $28.35_{\pm 0.52}$ \\
GDUMB \citep{prabhu2020gdumb} & 4k  & $54.12_{\pm 1.10}$ & $14.04_{\pm 0.86}$ \\
iCaRL \citep{rebuffi2017icarl} & 4k  & $25.81_{\pm 0.97}$ & $5.79_{\pm 0.16}$  \\
LUCIR \citep{hou2019learning} & 4k  & $30.15_{\pm 1.45}$ & $19.17_{\pm 1.39}$ \\
DER++ \citep{buzzega2020dark} & 4k  & $57.90_{\pm 3.28}$ & $27.25_{\pm 2.17}$ \\
LODE \citep{liang2023loss}  & 4k  & $\mathbf{63.98}_{\pm 0.96}$ & $32.94_{\pm 1.06}$ \\
MANGO (ours) & 4k & $54.00_{\pm 1.60}$ & $\mathbf{33.01}_{\pm 0.96}$ \\
\bottomrule
\end{tabular}
\end{table}

\section{Backward Transfer in CIL Settings}
\label{app:bwt}
Backward Transfer (BWT) measures impact of learning new samples on the performance of previously learned ones. In offline continual learning, it is a standard evaluation metric as the tasks can be evaluated again post training. However, in strict online settings of OCL where each sample is observed only once, BWT becomes less informative. Near-zero BWT  can be achieved by conservative updates that reduce CF but at the cost of plasticity. For example, in Table \ref{tab:bwt_cil} we can observe that ER-ACE achieves near-zero BWT but at the cost of accuracy.  Considering this, we follow standard benchmark OCL metrics like Acc, AAA, WC-Acc instead, to promote fair evaluation. 
Table \ref{tab:bwt_cil} reports BWT of all baselines on CIFAR-100 and Tiny-ImageNet on replay buffer sizes of 2000 and 4000. 
\begin{equation}
\text{BWT} = \frac{1}{T-1} \sum{j=1}^{T-1} \left( \mathbf{A}{T,j} - \mathbf{A}{j,j} \right)
\end{equation}

\begin{table}[h]
\caption{Backward Transfer (BWT) on CIFAR-100 and Tiny-ImageNet, single-pass over 5 seeds.}
\label{tab:bwt_cil}
\centering
\small
\setlength{\tabcolsep}{4pt}
\begin{tabular}{lccc}
\toprule
Method & Buf. & \textbf{CIFAR-100} (\%) & \textbf{Tiny-ImageNet} (\%) \\
\midrule
\multicolumn{4}{l}{\textit{Buffer $= 2{,}000$}} \\[2pt]
FT           & ---  & $-31.27_{\pm 1.58}$ & $-14.04_{\pm 1.23}$ \\
ER \citep{chaudhry2019continual}          & 2k   & $-46.91_{\pm 1.75}$ & $-30.76_{\pm 0.50}$ \\
ER-ACE \citep{caccia2022new}      & 2k   & $\mathbf{+0.51}_{\pm 1.38}$ & $-30.76_{\pm 0.45}$ \\
GDUMB \citep{prabhu2020gdumb}      & 2k   & $-12.94_{\pm 0.83}$ & $-4.14_{\pm 0.23}$  \\
iCaRL \citep{rebuffi2017icarl}  & 2k   & $-3.88_{\pm 0.74}$  & $\mathbf{-1.43}_{\pm 0.20}$ \\
LUCIR \citep{hou2019learning}      & 2k   & $-10.22_{\pm 0.68}$ & $-22.49_{\pm 1.15}$ \\
DER++ \citep{buzzega2020dark}        & 2k   & $-54.46_{\pm 1.10}$ & $-26.88_{\pm 1.35}$ \\
LODE  \citep{liang2023loss}        & 2k   & $-8.95_{\pm 2.11}$  & $-16.72_{\pm 1.64}$ \\
MANGO (ours) & 2k   & $-38.17_{\pm 1.12}$ & $-6.54_{\pm 0.91}$  \\
\midrule
\multicolumn{4}{l}{\textit{Buffer $= 4{,}000$}} \\[2pt]
ER  \citep{chaudhry2019continual}   & 4k   & $-41.51_{\pm 1.67}$ & $-28.47_{\pm 0.49}$ \\
ER-ACE \citep{caccia2022new}   & 4k   & $\mathbf{+0.38}_{\pm 2.10}$ & $-28.47_{\pm 0.48}$ \\
GDUMB \citep{prabhu2020gdumb}   & 4k   & $-13.22_{\pm 1.45}$ & $-6.03_{\pm 0.50}$  \\
iCaRL \citep{rebuffi2017icarl}  & 4k   & $-4.85_{\pm 0.48}$  & $\mathbf{-1.89}_{\pm 0.18}$ \\
LUCIR \citep{hou2019learning}         & 4k   & $-10.58_{\pm 1.74}$ & $-21.72_{\pm 0.83}$ \\
DER++ \citep{buzzega2020dark}   & 4k   & $-57.09_{\pm 1.79}$ & $-22.54_{\pm 1.83}$ \\
LODE \citep{liang2023loss}      & 4k   & $-2.48_{\pm 2.88}$  & $-11.67_{\pm 2.07}$ \\
MANGO (ours) & 4k   & $-35.70_{\pm 2.70}$ & $-6.60_{\pm 0.93}$  \\
\bottomrule
\end{tabular}
\end{table}

\section{Implementation Details}
\label{app:impl}

\subsection{Architecture Configurations}
\label{app:arch}

Experiments have been performed on ResNet-18 \citep{he2016deep} architecture. CIFAR-100 is divided into 20 disjoint tasks, each containing 5 classes and 2.5k training samples per task. Tiny-ImageNet is split into 10 disjoint tasks with 20 classes per task and 10k training samples per task. CLEAR-10 consists of 10 DIL tasks with each task containing 10 classes and 3k training samples. For CIFAR-100 dataset which has ($32{\times}32$) inputs, initial layers use a $3{\times}3$ kernel,  the max-pooling layer is removed for low-resolution input conditions. For Tiny-ImageNet which has ($64{\times}64$) inputs and CLEAR-10 datasets, standard ResNet-18 has been used for training. For CIL, all models share a classification head across tasks. TIL evaluation, separate task head are used for each tasks. No pre-trained or feature extraction is performed. All models are trained using random initialization under strict online settings. The layer-wise regularization coefficients are grouped on ethe following components of the architecture: Initial convolutional block, layer 1, layer 2, layer 3 and layer 4 along with the final classifier head, producing 6 scalar lambdas.  $\lambda$ values are parameterized in log-space, then exponentiated to be used in the regularization term. This ensures that $\lambda$  strictly stays positive during training, avoiding constrained optimization. They are then initialized at the start of training.

\subsection{Hyperparameters}
\label{app:hparams}

\begin{table}[h]
\caption{Hyperparameters used in baseline run of all.}
\label{tab:hyperparams}
\centering
\small
\begin{tabular}{lccc}
\toprule
\textbf{Hyperparameter} & \textbf{CIFAR-100} & \textbf{Tiny-ImageNet} & \textbf{CLEAR-10} \\
\midrule
Model learning rate $\eta$ & 0.02 & 0.05 & 0.05 \\
Lambda learning rate $\eta_\lambda$ & $2\times10^{-3}$ & $1\times10^{-3}$ & $2\times10^{-3}$ \\
Lambda initialization (log-space) & -7.6 & -7.6 & -7.6 \\
Number of glances $g$ & 3 & 3 & 3 \\
Online batch size & 32 & 32 & 32 \\
Replay batch size & 64 & 64 & 64 \\
SGD momentum & 0.9 & 0.9 & 0.9 \\
SGD weight decay & 0.0 & 0.0 & 0.0 \\
Buffer sizes evaluated & \{1k,2k,4k\}& \{2k,4k\}& \{2k,4k\}\\
Random seeds & 5 & 5 & 5 \\
$\lambda$ grouping & 5 groups & 5 groups & 5 groups \\
Meta-update frequency & Every 3 batches & Every 5 batches & Every 3 batches \\
Meta batch size & 256 & 256 & 256 \\
Optimizer & SGD & SGD & SGD \\
Lambda optimizer & Adam & Adam & Adam \\
\bottomrule
\end{tabular}
\end{table}

The model is trained in a strict-online setting, where tasks arrive sequentially and each sample is only seen once. Data arrives in mini-batches and are processed in three glances to improve stability, with three consecutive forward and backward passes over the batch. Online and replay batch sizes are fixed to 32 and 64 respectively, ensuring fair comparison among baselines. Stochastic gradient descent (SGD) is used for optimizing the model with 0.9 momentum and no weight decay. The stability coefficients use gradient descent with a learning rate of $\eta_\lambda$ (see Table~\ref{tab:hyperparams}).  All training is run across 5 random seeds. Baseline methods are run on model learning rate used in MANGO across datasets for fair comparison.Table \ref{tab:distill_hparams} contains distillation hyperparameters used for running baselines which work on distillation based models \citep{buzzega2020dark,liang2023loss,hou2019learning,rebuffi2017icarl}.

\begin{table}[h]
\caption{Hyperparameters used for distillation-based baseline methods.}
\label{tab:distill_hparams}
\centering
\small
\begin{tabular}{lcc}
\toprule
\textbf{Method} & \textbf{Hyperparameter} & \textbf{Value} \\
\midrule
DER++ \citep{buzzega2020dark}  & Distillation weight $\alpha$ & 0.2 \\
DER++ \citep{buzzega2020dark} & Replay CE weight $\beta$ & 0.5 \\
LUCIR \citep{hou2019learning} & Cosine margin $m$ & 0.5 \\
LUCIR \citep{hou2019learning} & Negative exemplars $K$ & 2 \\
LODE  \citep{liang2023loss}  & Distillation objective & Logit-based \\
iCaRL iCaRL \citep{rebuffi2017icarl} & Classifier type & Nearest-mean exemplars \\
\bottomrule
\end{tabular}
\end{table}

\subsection{Replay Memory Buffer Management}
\label{app:buffer}
The replay memory buffer has been maintained using reservoir sampling \citep{vitter1985random}. This ensures that there are equal chances for each past sample to be present in the buffer, irrespective of their processing order. Reservoir sampling is performed to produce unbiased samples on the data seen so far.  After each step update in the model, the buffer is updated by replacing randomly chosen existing replay samples with incoming samples. Meta-learned regularization step uses replay samples for computing $\mathcal{L}_\text{meta}$ to update $\lambda$. Old parameters  $\theta^\text{old}$ in the regularization term are updated to the present state of the model at the end of each task. 

\section{Extended Ablation Study Analysis}
\label{app:ablation}
\begin{table}[h]
\caption{Ablation study at $M{=}4000$, single-pass OCL over 5
seeds.}
\label{tab:ablation_4k}
\centering
\small
\setlength{\tabcolsep}{4pt}
\begin{tabular}{lcccc}
\toprule
& & \multicolumn{1}{c}{\textbf{CIFAR-100}}
  & \multicolumn{1}{c}{\textbf{Tiny-ImageNet}}
  & \multicolumn{1}{c}{\textbf{CLEAR-10}} \\
\cmidrule(lr){3-3}\cmidrule(lr){4-4}\cmidrule(lr){5-5}
Method & Buf. & CIL Acc (\%) & CIL Acc (\%) & DIL Acc (\%) \\
\midrule
MANGO                 & 4k & $\mathbf{20.64}_{\pm 1.15}$ & $\mathbf{24.73}_{\pm 0.80}$ & $\mathbf{67.67}_{\pm 1.94}$ \\
w/o Replay (FT)       & ---& $1.65_{\pm 0.29}$           & $1.94_{\pm 0.40}$           & $62.99_{\pm 0.64}$ \\
w/o Meta-$\lambda$    & 4k & $20.16_{\pm 1.16}$ & $19.29_{\pm 0.36}$       & $67.10_{\pm 2.01}$ \\
w/ only Replay (ER)   & 4k & $12.32_{\pm 0.42}$          & $5.93_{\pm 0.34}$           & $67.10_{\pm 1.95}$ \\
w/o Regularization    & 4k & $18.69_{\pm 1.87}$          & $20.27_{\pm 0.89}$ & $65.37_{\pm 0.28}$ \\
\bottomrule
\end{tabular}
\end{table}
Table~\ref{tab:ablation_4k} reports the extended ablation study of MANGO at an additional buffer size of $4000$, adding to the main paper ablation done on buffer size of $2000$, shown in Table \ref{app:ablation} . 
Results on buffer size 4000 are identical and follow the comparable analysis as section~\ref{sec:ablation}. 

\newpage

\section*{NeurIPS Paper Checklist}

\begin{enumerate}

\item {\bf Claims}
    \item[] Question: Do the main claims made in the abstract and introduction accurately reflect the paper's contributions and scope?
    \item[] Answer: \answerYes{} 
    \item[] Justification: \justification{All claims made in the abstract and introduction are experimentally supported over 3 datasets, 5 seeds and 2 buffer sizes.}

\item {\bf Limitations}
    \item[] Question: Does the paper discuss the limitations of the work performed by the authors?
    \item[] Answer: \answerYes{} 
    \item[] Justification: \justification{Our paper discusses the limitations of our proposed method.}

\item {\bf Theory assumptions and proofs}
    \item[] Question: For each theoretical result, does the paper provide the full set of assumptions and a complete (and correct) proof?
    \item[] Answer: \answerNA{} 
    \item[] Justification: \justification{Our approach is an empirical method.}

    \item {\bf Experimental result reproducibility}
    \item[] Question: Does the paper fully disclose all the information needed to reproduce the main experimental results of the paper to the extent that it affects the main claims and/or conclusions of the paper (regardless of whether the code and data are provided or not)?
    \item[] Answer: \answerYes{} 
    \item[] Justification: \justification{The paper discloses all necessary experimental settings to reproduce our algorithm.}

\item {\bf Open access to data and code}
    \item[] Question: Does the paper provide open access to the data and code, with sufficient instructions to faithfully reproduce the main experimental results, as described in supplemental material?
    \item[] Answer: \answerYes{} 
    \item[] Justification: \justification{The code is provided as a .zip file.}

\item {\bf Experimental setting/details}
    \item[] Question: Does the paper specify all the training and test details (e.g., data splits, hyperparameters, how they were chosen, type of optimizer) necessary to understand the results?
    \item[] Answer: \answerYes{} 
    \item[] Justification: \justification{All information regarding all hyperparameters, datasets, baselines and evaluation protocol is discussed in the paper.}

\item {\bf Experiment statistical significance}
    \item[] Question: Does the paper report error bars suitably and correctly defined or other appropriate information about the statistical significance of the experiments?
    \item[] Answer: \answerYes{} 
    \item[] Justification: \justification{The paper contains results over random seeds and buffers.}

\item {\bf Experiments compute resources}
    \item[] Question: For each experiment, does the paper provide sufficient information on the computer resources (type of compute workers, memory, time of execution) needed to reproduce the experiments?
    \item[] Answer: \answerYes{} 
    \item[] Justification: \justification{Computational resources used for experimentation is mentioned in the paper.}
    
\item {\bf Code of ethics}
    \item[] Question: Does the research conducted in the paper conform, in every respect, with the NeurIPS Code of Ethics \url{https://neurips.cc/public/EthicsGuidelines}?
    \item[] Answer: \answerYes{} 
    \item[] Justification: \justification{Our research follows all codes of ethics set by NeurIPS.}

\item {\bf Broader impacts}
    \item[] Question: Does the paper discuss both potential positive societal impacts and negative societal impacts of the work performed?
    \item[] Answer: \answerYes{} 
    \item[] Justification: \justification{The paper discusses a single epoch machine learning method to reduce computations in training models.}
    
\item {\bf Safeguards}
    \item[] Question: Does the paper describe safeguards that have been put in place for responsible release of data or models that have a high risk for misuse (e.g., pre-trained language models, image generators, or scraped datasets)?
    \item[] Answer: \answerNA{} 
    \item[] Justification: \justification{The paper doesn't pose any such risks.}

\item {\bf Licenses for existing assets}
    \item[] Question: Are the creators or original owners of assets (e.g., code, data, models), used in the paper, properly credited and are the license and terms of use explicitly mentioned and properly respected?
    \item[] Answer: \answerYes{} 
    \item[] Justification: \justification{CIFAR-100, Tiny-ImageNet and CLEAR-10 are standard public benchmark datasets.}

\item {\bf New assets}
    \item[] Question: Are new assets introduced in the paper well documented and is the documentation provided alongside the assets?
    \item[] Answer: \answerNA{} 
    \item[] Justification: \justification{This paper does not introduce new datasets, benchmarks, or externally released assets. We evaluate on existing public continual learning benchmarks and provide implementation details within the paper.}

\item {\bf Crowdsourcing and research with human subjects}
    \item[] Question: For crowdsourcing experiments and research with human subjects, does the paper include the full text of instructions given to participants and screenshots, if applicable, as well as details about compensation (if any)? 
    \item[] Answer: \answerNA{} 
    \item[] Justification: \justification{The paper doesn't involve crowdsourcing or human experiments.}

\item {\bf Institutional review board (IRB) approvals or equivalent for research with human subjects}
    \item[] Question: Does the paper describe potential risks incurred by study participants, whether such risks were disclosed to the subjects, and whether Institutional Review Board (IRB) approvals (or an equivalent approval/review based on the requirements of your country or institution) were obtained?
    \item[] Answer: \answerNA{} 
    \item[] Justification: \justification{The paper doesn't involve crowdsourcing or human experiments.}

\item {\bf Declaration of LLM usage}
    \item[] Question: Does the paper describe the usage of LLMs if it is an important, original, or non-standard component of the core methods in this research? Note that if the LLM is used only for writing, editing, or formatting purposes and does \emph{not} impact the core methodology, scientific rigor, or originality of the research, declaration is not required.
    \item[] Answer: \answerNA{} 
    \item[] Justification: \justification{The paper does not involve LLMs in important, original or non-standard components.}

\end{enumerate}

\end{document}